\documentclass[sigconf]{acmart}
\usepackage{array}
\usepackage{tabularx}
\usepackage{graphicx}
\AtBeginDocument{%
  }

\setcopyright{acmlicensed}
\copyrightyear{2018}
\acmYear{2018}
\acmDOI{XXXXXXX.XXXXXXX}
\acmConference[Conference acronym 'XX]{Make sure to enter the correct
  conference title from your rights confirmation email}{June 03--05,
  2018}{Woodstock, NY}
\acmISBN{978-1-4503-XXXX-X/2018/06}

\begin{document}

\title{Mira-Embeddings-V1: Domain-Adapted Semantic Reranking for Recruitment via LLM-Synthesized Data}



\author{Zhaohua Liang}
\affiliation{
  \institution{OpenJobs AI}
  \city{San Francisico, CA}
  \country{USA}}
\email{kokoo1735172984@gmail.com}

\author{Zhilin Wang}
\authornote{Corresponding author.}
\affiliation{
  \institution{OpenJobs AI}
  \city{San Francisico, CA}
  \country{USA}}
\email{jerry.wang@openjobs-ai.com}

\author{Renjie Cao}
\affiliation{
  \institution{OpenJobs AI}
  \city{San Francisico, CA}
  \country{USA}}
\email{heymegumi06@gmail.com}

\author{Yining Zhang}
\affiliation{
  \institution{OpenJobs AI}
  \city{San Francisico, CA}
  \country{USA}}
\email{zhangyiningfudan@gmail.com}

\renewcommand{\shortauthors}{Trovato et al.}

\begin{abstract}
Candidate sourcing for recruiters is best viewed as a two stage re-
trieval and reranking pipeline with recall as the primary objective
under a limited review budget. In the primary setting studied in this paper, an upstream production retriever first returns a candidate shortlist for each job description (JD), and our goal is to rerank that shortlist so that qualified candidates appear as high as possible. We present mira-embeddings-v1, a semantic reranking system for
the recruitment domain that reshapes the embedding space with
large language model (LLM)-synthesized training data and further
corrects boundary confusions with a lightweight reranking head. Starting from real JDs, we build a five stage prompt pipeline to generate diverse positive and hard negative samples that sculpt the model's semantic space from multiple angles. We then apply a two round LoRA adaptation: JD--JD contrastive training followed by JD--CV triplet alignment on a heterogeneous text JD--CV dataset. Importantly, these gains are pursued without large scale manually labeled industrial training pairs: a modest set of real JDs is expanded into supervision at the task level through LLM synthesis. Finally, a BoundaryHead multi-layer perceptron (MLP) reranks the Top-$K$ results to further distinguish between roles that share the same title but differ in scope. On a local pool of 300 real JDs with candidates returned by an upstream production retriever, mira-embeddings-v1 improves Recall@50 from 68.89\% (baseline) to 77.55\% while lifting Precision@10 from 35.77\% to 39.62\%. On a supportive global pool protocol over 44,138 candidates judged by a Qwen3-32B rubric, it achieves Recall@200 of 0.7047 versus 0.5969 for the baseline. These results show that supervision synthesized by LLMs with data efficiency, plus reranking with boundary awareness, yields robust gains without introducing a heavy cross encoder in our current pipeline.
\end{abstract}

\received{20 February 2007}
\received[revised]{12 March 2009}
\received[accepted]{5 June 2009}

\maketitle

\section{Introduction}

\label{sec:intro}

 In candidate sourcing for recruiters, an upstream production retriever first returns a candidate shortlist for each job description (JD), and the central ranking problem is how to reorder that shortlist so that qualified candidates
  appear within the positions recruiters can realistically inspect. In this setting, the main operational risk is
not showing one extra partially matching candidate, but missing a
truly qualified candidate who then receives no exposure at all. For
this reason, JD to CV shortlist reranking should be treated as a recall
oriented recommendation problem under a fixed review budget.
Because JDs specify future role requirements while CVs summarize
past experience, the task is also a heterogeneous retrieval problem
with a systematic view gap.

Existing methods for retrieval and recruitment matching improve general semantic relevance~\cite{reimers2019sentencebert,karpukhin2020dpr,bian2019domain,jiang2020featurefusion,bian2020mvcn,fang2023recruitpro,yu2024confit,yu2025confitv2,gao2023hyde}. Dense retrieval with dual encoders is attractive because it supports embeddings that are reusable and efficient search for nearest neighbors, while prior work in recruitment has explored matching methods adapted to the domain and fine tuning based on synthetic data. However, these methods often fail in the cases that matter most in shortlist generation: candidates who are near misses with similar titles or overlapping skills yet incorrect ownership level, functional scope, seniority, or mandatory constraints. These errors at the role boundary are costly in settings where recall is the primary objective because they occupy scarce shortlist positions and crowd out truly qualified candidates.

We therefore treat \emph{role-boundary confusion} as an explicit training target rather than assuming that a stronger general encoder alone will solve it. More concretely, we aim to teach the system not only what a relevant JD--CV match looks like, but also which near-miss cases should be separated: semantic paraphrases that should remain close, boundary mismatches that should be pushed apart, and hard constraint conflicts that should be demoted despite topical overlap. Starting from a modest set of real JDs, we use positives and hard negatives generated by LLMs to create supervision targeted at specific tasks at low marginal cost~\cite{yu2024confit,yu2025confitv2,gao2023hyde}. On this basis, we propose \textbf{mira-embeddings-v1}: a five-stage prompt pipeline generates structured JD--JD supervision; two rounds of LoRA adaptation~\cite{hu2021lora} learn recruitment semantics and then JD--CV alignment; and a lightweight BoundaryHead makes local Top-$K$ corrections without introducing a heavy cross encoder.

Experiments on a local pool (300 JDs with candidates from an upstream production retriever) and a supportive larger pool (44K candidates with LLM judgments based on a rubric) show a consistent pattern: the domain-adapted encoder supplies the main recall gain, and BoundaryHead adds smaller residual improvements in the shortlist region. Evaluation on the local pool is the primary evidence because it matches the actual recruiter review scenario, while the larger pool protocol serves as a supportive consistency check. Together, these results support a focused claim: explicit boundary-aware supervision improves shortlist recall for recruiters at modest data construction cost.

Our main contributions are:
\begin{itemize}
  \item We identify candidate sourcing for recruiters as a reranking problem in which recall is the primary objective and near misses involving role boundaries or hard constraints are a central source of error.
  \item We build an LLM supervision pipeline with five stages at low cost that expands real JDs into structured positives and hard negatives for paraphrase robustness, role boundary separation, and constraint sensitivity.
  \item We introduce a two round adaptation pipeline, with JD--JD contrastive tuning followed by JD--CV triplet alignment. This provides the main source of recall improvement.
  \item We introduce a lightweight \textbf{BoundaryHead} that provides residual local correction for boundary-confused Top-$K$ candidates.
  \item We establish an evaluation oriented toward recall across a primary Local Pool and a supportive Global Pool, showing that encoder adaptation drives most gains while BoundaryHead adds further shortlist improvements.
\end{itemize}

\section{Related Work}
\label{sec:related}

\paragraph{Recruitment modeling: pair scoring is not shortlist retrieval.}
Prior work on person--job fit has mostly framed recruitment as a \emph{pairwise} matching problem: given one JD and one CV, predict whether the pair is compatible. Early neural approaches learn joint JD--CV representations using ability-aware hierarchical attention or bipartite CNN architectures~\cite{qin2018pjfnn,zhu2018pjf}, and later work strengthens this formulation through topic-based ability modeling and improved optimization objectives~\cite{qin2020enhanced}. Other studies enrich the available signals, for example by incorporating interview records~\cite{shen2018joint} or modeling recruiter preference and historical interview behavior with profiling-memory mechanisms~\cite{yan2019interview}. Taken together, these studies show that recruitment benefits from domain-aware modeling, but they still target fit estimation for individual pairs. That leaves open the retrieval question studied here: under a fixed recruiter review budget, how should the embedding space be organized so that strong candidates can be recovered at high recall, especially when many near-miss candidates are semantically similar but differ on role boundaries or hard constraints such as required certifications or minimum experience?

\paragraph{General dense retrieval and synthetic supervision improve relevance, but not the failure mode we target.}
Dense retrieval has made shared query--document embeddings a standard approach for efficient nearest-neighbor search, supported by unsupervised contrastive pretraining~\cite{izacard2022contriever}, instruction-tuned embeddings~\cite{su2023instructor,ni2022sentencet5}, and large-scale weak supervision~\cite{wang2022e5}. This line of work has produced strong general-purpose encoders such as E5~\cite{wang2022e5}, GTE~\cite{li2023gte}, BGE~\cite{xiao2023cpack}, BGE-M3~\cite{chen2024bgem3}, and Jina Embeddings v3~\cite{sturua2024jinav3}. At the same time, parameter-efficient adaptation methods such as LoRA and QLoRA have made downstream specialization practical even with limited computational resources~\cite{hu2021lora,dettmers2023qlora}. In parallel, synthetic-data approaches show that LLM-generated supervision can substantially improve retrieval. InPars and InPars-v2 synthesize queries from documents~\cite{bonifacio2022inpars,jeronymo2023inparsv2}, Promptagator bootstraps training data from a small seed set~\cite{dai2023promptagator}, HyDE uses hypothetical documents for zero-shot retrieval~\cite{gao2023hyde}, and recent work argues that diverse synthetic tasks can strengthen embedding models more broadly~\cite{wang2024improving}. What is still missing for our setting is not a stronger generic relevance recipe, but a formulation centered on recruitment-specific shortlist errors. Existing adaptation and synthesis methods improve semantic relevance in a broad sense, yet they do not explicitly treat \emph{role-boundary confusion}, namely near-miss candidates that are topically close but wrong on seniority, function, domain, or mandatory requirements, as the failure mode to be diagnosed and suppressed.

\paragraph{Heavy reranking is effective, but our question is whether targeted local correction is enough.}
A separate line of work improves retrieval with powerful second-stage rerankers. Cross-encoders jointly score query--document pairs and are highly effective, but their computational cost scales linearly with the number of candidates~\cite{nogueira2019passage,nogueira2020document}. Late-interaction models such as ColBERT offer a more efficient compromise through precomputed token-level representations~\cite{khattab2020colbert}, and sequence-to-sequence or LLM-based rerankers further improve generic ranking quality~\cite{nogueira2020document,sun2023rankgpt,pradeep2023rankzephyr}. These methods are strong when the objective is broad relevance optimization with a heavy reranking stage. Our setting is narrower: after first-stage retrieval, can residual recruitment-specific near-miss errors be corrected with a cheaper, more targeted mechanism that does not require full pairwise rescoring? This is the setting in which we position the paper. Our comparison is therefore against strong general-purpose embedding backbones under the retrieval setup studied here, rather than a claim of state-of-the-art performance across the full recruitment literature. The next section formalizes this narrower task, its review-budget assumption, and the design requirements implied by role-boundary confusion.

\section{Problem Formulation}
\label{sec:problem}

\subsection{Task Definition}

Let $q$ denote a job description (JD), and let
$\mathcal{C}(q) = \{c_1, c_2, \ldots, c_M\}$ denote the candidate pool
returned for $q$ by an upstream retrieval system in the primary
setting studied in this paper. We learn a $d$-dimensional
embedding function $f\!: \mathcal{X} \to \mathbb{R}^d$ such that
semantically relevant JD--CV pairs within this pool are mapped close
together in the embedding space:
\begin{equation}
  \text{sim}(f(q),\, f(c^+)) \;\gg\; \text{sim}(f(q),\, f(c^-)),
  \label{eq:objective}
\end{equation}
where $c^+$ is a ground-truth matching CV in $\mathcal{C}(q)$ and
$c^-$ is a non-matching one, and $\text{sim}(\cdot,\cdot)$ denotes
cosine similarity. Our primary objective is to rank candidates within
$\mathcal{C}(q)$ so that truly qualified candidates appear as high as
possible within the recruiter's review budget. The paper's main claim is therefore about reranking quality within an upstream candidate pool. Unlike symmetric retrieval tasks,
JD-to-CV matching is
\emph{inherently asymmetric}: JDs describe future role requirements in imperative
language, while CVs narrate actual past experience, producing
systematic surface-form gaps that general-purpose encoders tend to
conflate.

\subsection{Recall-First Evaluation}

Given a JD $q$, let $\mathcal{R}_K(q) \subset \mathcal{C}(q)$ denote
the top-$K$ candidates returned after the encoder ranking, and let
$\mathcal{P}(q) \subset \mathcal{C}(q)$ be the set of ground-truth
positive candidates. We evaluate retrieval quality using:
\begin{align}
  \text{Recall@}K   &= \frac{|\mathcal{R}_K(q) \cap \mathcal{P}(q)|}{|\mathcal{P}(q)|},
  \label{eq:recall} \\[2pt]
  \text{Precision@}K &= \frac{|\mathcal{R}_K(q) \cap \mathcal{P}(q)|}{K}.
  \label{eq:precision}
\end{align}
We treat Recall@$K$ as the primary metric. In recruiter-facing
candidate sourcing, the cost of a false negative (missing a qualified
candidate) outweighs the cost of a false positive (surfacing an
unqualified one that a recruiter will filter). Precision@$K$ serves as
a secondary indicator to guard against excessive recall-precision
trade-offs.

We report results on two complementary test pools (detailed in
Section~\ref{sec:eval}): a \emph{Local Pool} that serves as the
primary production reranking setting (300 JDs, ${\sim}200$ initial
candidates each), and a \emph{supportive Global Pool} protocol that
probes whether the same directional conclusions remain under broader
direct retrieval (44,138 candidates, LLM-as-Judge ground truth).

\subsection{Two-Stage Ranking Framework}

The overall system decomposes ranking into two stages.

\textbf{Stage 1 — Encoder Scoring.} Given JD $q$, we compute cosine
similarity between the query and each candidate in the returned pool
$\mathcal{C}(q)$ using the fine-tuned encoder $f$:
\begin{equation}
  s_{\cos}(q, c) = \text{sim}(f(q),\, f(c)),
  \quad c \in \mathcal{C}(q).
  \label{eq:stage1}
\end{equation}

\textbf{Stage 2 — Boundary Reranking.} A lightweight
\textbf{BoundaryHead} MLP scores each candidate in $\mathcal{C}(q)$
for role-boundary conformance. The final ranking score is:
\begin{equation}
  s_{\text{final}}(q, c) =
    s_{\cos}(q, c) \;-\; \lambda \cdot s_{\text{boundary}}(q, c),
  \label{eq:stage2}
\end{equation}
where $s_{\text{boundary}} \in [0,1]$ is the BoundaryHead output
(higher means stronger role-boundary mismatch), and $\lambda$ is a
small weighting coefficient tuned on the validation set. In the primary
Local Pool setting, Stage~1 is fixed by the upstream production
retriever, and our method is evaluated as a reranker over the returned
shortlist. In the supportive Global Pool protocol, we instantiate a
broader first-stage retrieval step explicitly and set $K_1{=}1000$,
with $\lambda{=}0.1$ for BoundaryHead reranking.
Section~\ref{sec:eval} describes these evaluation settings in detail.

\section{Method}
\label{sec:method}

Our goal is to improve recruiter-facing, recall-first shortlist reranking under limited supervision by addressing the specific failure modes that are central to this task setting: semantically similar but recruiter-invalid near misses, especially role-boundary confusions and hard-constraint violations. This motivates to three method requirements. First, supervision should not only mark what constitutes a relevant match, but also distinguish whether a near-miss case should remain close, be separated, or be locally demoted. Second, adaptation should bridge both recruitment-domain semantics and the heterogeneous JD--CV view gap without relying on large-scale human labels. Third, any reranking component should serve as a conservative local correction rather than replacing the main retrieval model.

We build \textbf{mira-embeddings-v1} in three stages, illustrated in
Figure~\ref{fig:overview}: (\textit{i})~a five-stage LLM-driven prompt
pipeline that generates structured training pairs from real JDs together with a
heterogeneous text JD--CV dataset; (\textit{ii})~two rounds of LoRA fine-tuning
that progressively adapt a general-purpose encoder to recruitment-domain
matching; and (\textit{iii})~a lightweight BoundaryHead MLP trained on
top of the frozen encoder to suppress role-boundary confusions at
reranking time.

\begin{figure*}[htbp]
    \centering
    \includegraphics[width=0.8\textwidth]{Figure1.png}
    \caption{Overview of the mira-embeddings-v1 ranking system. Top: an LLM-driven prompt pipeline generates JD--JD pairs, a JD--CV synthetic dataset, and role-boundary pairs from raw JDs. Middle: training consists of two rounds of LoRA fine-tuning, followed by a BoundaryHead MLP trained on the frozen encoder to reduce role-boundary confusions. Bottom: at inference time, a query JD scores candidates in the returned pool with the fine-tuned encoder and reranks them with the BoundaryHead. Light green arrows denote the flow of generated data into training.}
    \label{fig:overview}
\end{figure*}

\subsection{Supervision Design for the Target Failure Modes}
\label{sec:data}

\subsubsection{Five-Stage JD Prompt Pipeline}
\label{sec:pipeline}

The JD-only supervision is designed to teach the encoder three behaviors that standard similarity training does not cleanly isolate: robustness to benign phrasing variation, separation of same-title but different-scope roles, and sensitivity to hard-constraint conflicts. To keep all generated pairs anchored to real demand, every sample must remain programmatically grounded in a real JD and pass script-level validation before entering training.

\begin{table}[t]
  \caption{Five-stage JD prompt pipeline. Prompt~2, Prompt~4, and
           Prompt~5 consume the Prompt~1 structured record directly,
           while Prompt~3 generates a summary-to-card positive from the
           raw JD summary. Pair counts report retained training pairs
           after script-level validation and deduplication.}
  \label{tab:pipeline}
  \small
  \begin{tabular}{clp{3.1cm}r}
    \toprule
    Stage & Type & Training Target & Pairs \\
    \midrule
    1 & Extraction & Structured anchor (title, skills, constraints) & 3,317 \\
    2 & pos & Paraphrase robustness (synonym/abbreviation rewrites) & 3,317 \\
    3 & pos & Style-noise reduction (summary-to-card anchor) & 5,342 \\
    4 & hard neg & Role-boundary: same title, shallower ownership/scope & 9,951 \\
    5 & hard neg & Constraint conflict: single non-location field flipped & 6,634 \\
    \midrule
    \multicolumn{3}{l}{\textit{Total (3,317 source JDs)}} & 28,561 \\
    \bottomrule
  \end{tabular}
\end{table}

Prompt~1 extracts a conservative structured anchor from each real JD. Prompts~2--3 then generate positive pairs for paraphrase robustness and style normalization, while Prompts~4--5 generate hard negatives for role-boundary confusion and hard-constraint conflicts. After expansion, validation, and deduplication, the 3,317 source JDs yield 28,561 usable JD--JD pairs (Table~\ref{tab:pipeline}). Cleaned and condensed prompt templates are provided in the supplementary material.

\subsubsection{JD--CV Synthetic Data Construction}
\label{sec:jdcv-data}

The second supervision need is heterogeneous alignment. After learning recruitment-specific semantics on JD--JD data, the model must also learn how those semantics transfer from a job description to a CV. We therefore construct the JD--CV corpus from two complementary synthetic sources: a pan-domain balanced generator that covers diverse roles with clear positives and screening-style negatives, and a LinkedIn-style hard-negative generator that emphasizes deceptive high-overlap CV cases such as ambiguity, inflation, contractor/intern confusion, career pivots, influencer noise, and skill stuffing. Both sources are schema-validated and normalized into a unified JSONL format, yielding the corpus used for Round~2 adaptation.

\subsection{Progressive Adaptation via Two Rounds of LoRA Fine-Tuning}
\label{sec:lora}

The encoder must solve two distinct adaptation problems. It first needs to internalize recruitment-domain semantics, and then bridge the heterogeneous semantic gap between JDs and CVs. We therefore adapt Jina Embeddings~v3 in two sequential LoRA rounds rather than a single joint pass. The first round uses JD-only supervision to shape embedding geometry around recruitment-relevant similarity and separation; the second aligns JD queries with CV candidates on top of an already domain-aware representation. This ordering reduces the burden of handling domain shift and cross-view mismatch at the same time.

\subsubsection{Round 1: JD--JD Domain Alignment (mira-embeddings-v1-R1)}
\label{sec:r1}

Round~1 fine-tunes Jina Embeddings~v3 on the JD--JD corpus from
Section~\ref{sec:pipeline}. Its role is to inject recruitment-domain semantics and teach the encoder that some high-overlap pairs should still be pushed apart. We use a three-stage curriculum with the same LoRA adapter throughout (rank $r{=}16$, $\alpha{=}32$, five target linear layers: \texttt{dense}, \texttt{out\_proj}, \texttt{fc1}, \texttt{fc2}) and a learning rate of $8{\times}10^{-6}$.

\textbf{Stage 1 — Card Warm-up.}
The model trains on Prompt-3 canonical-card pairs, pulling a
natural-language JD summary toward its compact semantic card and providing a coarse domain anchor before harder ranking pairs are introduced.

\textbf{Stage 2 — Semantic Alignment.} The model then trains on Prompt-2 semantic-equivalent pairs. Stages~1 and~2 use the same pairwise alignment objective:
\begin{equation}
  \mathcal{L}_{\text{pair}} =
  w_p \bigl(1 - \cos(q, c)\bigr)^2
  + w_d \Bigl(\operatorname{MSE}(e_q, e_q^{(t)})
  + \operatorname{MSE}(e_c, e_c^{(t)})\Bigr),
  \label{eq:r1-pair}
\end{equation}
where $e_q = f(q)$ and $e_c = f(c)$ denote the embeddings of the query
and candidate produced by the current encoder, and $e_q^{(t)}$ and
$e_c^{(t)}$ are the corresponding embeddings from the original Jina
v3 model.
\textbf{Stage 3 — Main Ranking.} 
The model then switches to grouped ranking with one semantic positive $c^+$ from Prompt~2, one role-boundary negative $c^-_r$ from Prompt~4, and one constraint negative $c^-_k$ from Prompt~5. The ranking objective is:
\begin{equation}
  \mathcal{L} = w_p\,\mathcal{L}_{\text{pos}}
              + w_r\,\mathcal{L}_{\text{role}}
              + w_k\,\mathcal{L}_{\text{cons}}
              + w_d\,\mathcal{L}_{\text{distill}},
  \label{eq:r1loss}
\end{equation}
where
\begin{equation}
  \mathcal{L}_{\text{pos}} = \bigl(1 - \cos(q, c^+)\bigr)^2,
  \label{eq:r1-pos}
\end{equation}
\begin{equation}
  \mathcal{L}_{\text{role}} =
  \max \bigl(0,\; m_r - [\cos(q, c^+) - \cos(q, c^-_r)]\bigr)^2,
  \label{eq:r1-role}
\end{equation}
\begin{equation}
  \mathcal{L}_{\text{cons}} =
  \max \bigl(0,\; m_k - [\cos(q, c^+) - \cos(q, c^-_k)]\bigr)^2,
  \label{eq:r1-cons}
\end{equation}
and
\begin{equation}
  \mathcal{L}_{\text{distill}} =
  \operatorname{MSE}(e_q, e_q^{(t)}) +
  \operatorname{MSE}(e_{c^+}, e_{c^+}^{(t)}).
  \label{eq:r1-distill}
\end{equation}
Loss weights are
$(w_p, w_r, w_k, w_d) = (1.0,\;0.25,\;0.10,\;0.05)$, reflecting that
role-boundary discrimination requires a stronger gradient signal than
constraint learning. The implementation uses margins
$m_r{=}0.10$ and $m_k{=}0.05$. After all three stages, the LoRA adapter
is merged into the base model to produce \textbf{mira-embeddings-v1-R1}.

\subsubsection{Round 2: JD--CV Triplet Alignment (mira-embeddings-v1-R2)}
\label{sec:r2}

Round~2 starts from mira-embeddings-v1-R1 and fine-tunes on the
36,173-sample JD--CV corpus from Section~\ref{sec:jdcv-data}. Its purpose is not to relearn recruitment semantics, but to align the already adapted representation across the JD and CV views.
Training examples are formatted as triplets $(q, c^+, c^-)$, and we
use \textbf{Multiple Negatives Ranking Loss}
(MNRL)~\cite{reimers2019sentencebert}, which treats every positive
from another example in the same batch as an additional in-batch
negative.  For a mini-batch $\mathcal{B}$, the loss for anchor $q_i$
is:
\begin{equation}
  \mathcal{L}_\text{MNRL} = -\frac{1}{|\mathcal{B}|}
    \sum_{i \in \mathcal{B}}
    \log \frac{\exp(s_{q_i c_i^+}/\tau)}
    {\displaystyle\sum_{j \in \mathcal{B}}\exp(s_{q_i c_j^+}/\tau)
     + \sum_{j \in \mathcal{B}}\exp(s_{q_i c_j^-}/\tau)},
  \label{eq:mnrl}
\end{equation}
where $\tau$ is the temperature parameter and the denominator
accumulates both in-batch positives from other examples and the
explicit hard negatives supplied in every triplet.  Compared with a
standard margin-based triplet loss, MNRL exploits the full batch as a
negative pool and is less sensitive to margin tuning, which matters
when the hard negatives from Section~\ref{sec:jdcv-data} already vary
widely in difficulty.

After merging the Round-2 adapter weights into the base model, we obtain
\mbox{\textbf{mira-embeddings-v1-R2}}, the final embedding encoder.  Its weights
are subsequently \emph{frozen}; all later training (BoundaryHead,
Section~\ref{sec:boundary}) is confined to a separate local correction head.
Table~\ref{tab:lora} summarizes the key hyperparameters of both rounds.

\begin{table}[t]
  \caption{LoRA fine-tuning hyperparameters for both rounds.}
  \label{tab:lora}
  \small
  \setlength{\tabcolsep}{4pt}
  \begin{tabularx}{\columnwidth}{@{}>{\raggedright\arraybackslash}p{0.22\columnwidth}XX@{}}
    \toprule
    & \textbf{Round 1 (R1)} & \textbf{Round 2 (R2)} \\
    \midrule
    Base model        & Jina v3          & mira-embeddings-v1-R1 \\
    Training pairs    & 28,561 (Stage 2--5 JD--JD)  & 36,173 (JD--CV)   \\
    Loss              & Pairwise cosine alignment + margin ranking + distillation & MNRL         \\
    Learning rate     & $8{\times}10^{-6}$ & $1{\times}10^{-5}$ \\
    LoRA rank $r$     & 16               & 8                 \\
    LoRA $\alpha$     & 32               & 32                \\
    Max seq.\ length  & 512              & 2048              \\
    Max epochs        & 5                & 5                 \\
    \bottomrule
  \end{tabularx}
\end{table}

\subsection{Lightweight Local Correction with BoundaryHead}
\label{sec:boundary}

Even after encoder adaptation, some shortlist errors remain highly local. A candidate may be topically similar and therefore receive a strong embedding score, yet still need to be demoted because the actual ownership scope is shallower or otherwise boundary-misaligned. We therefore add BoundaryHead as a lightweight reranking correction on top of the frozen \textbf{mira-embeddings-v1-R2} encoder. Its role is deliberately narrower than that of the encoder. Round~1 uses role-boundary negatives to shape the global embedding geometry, whereas BoundaryHead uses the same signal to make local binary decisions among already high-similarity pairs.

\subsubsection{Architecture}

BoundaryHead is a shallow MLP (\textbf{PairHead}) that takes the
JD and CV embeddings generated by \textbf{mira-embeddings-v1-R2}
(Section~\ref{sec:r2}) and outputs a scalar boundary-mismatch
score $s_{\text{boundary}} \in [0,1]$.  The input feature vector
concatenates four interaction signals:
\begin{equation}
  \mathbf{h} = \bigl[f(q) \;\|\; f(c) \;\|\;
    \lvert f(q){-}f(c)\rvert \;\|\; f(q){\odot}f(c)\bigr]
    \in \mathbb{R}^{4d},
  \label{eq:pairhead-feat}
\end{equation}
where $f(\cdot)$ denotes the frozen embedding function implemented by
\textbf{mira-embeddings-v1-R2}, $d$ is the encoder hidden dimension,
$|\cdot|$ denotes element-wise absolute difference, and $\odot$
denotes element-wise product. The PairHead MLP is:
\begin{equation}
  \begin{gathered}
    \mathbf{z} = \operatorname{Dropout}
    \bigl(\operatorname{ReLU}(W_1 \mathbf{h} + b_1),\, p{=}0.1\bigr), \\
    s_{\text{boundary}} = W_2 \mathbf{z} + b_2,
  \end{gathered}
  \label{eq:pairhead-mlp}
\end{equation}
with $W_1 \in \mathbb{R}^{256 \times 4d}$ and
$W_2 \in \mathbb{R}^{1 \times 256}$. The encoder weights are completely
frozen during this stage; only the 256-dimensional MLP head is trained.

\subsubsection{Training Data}

BoundaryHead is trained on a binary discrimination task constructed
from two sources in a 1:1 ratio.

\textbf{Positive pairs} ($y=1$, boundary mismatch) come from the
Stage-4 role-boundary hard negatives (Section~\ref{sec:pipeline}).
These are JD pairs where the anchor has deep ownership and the
generated variant has a shallower, supporting-role rewrite under the
same title.  In the context of BoundaryHead, label~1 means ``this
pair shows a role-boundary mismatch that should be penalized at
reranking time.''

\textbf{Negative pairs} ($y=0$, genuine match) are drawn from the
positive pool of the Stage-2/3 corpus, filtered to retain only records
where the query and candidate titles match (exact, contains, or token
overlap $\geq 0.7$) \emph{and} the extracted responsibility fields
have a bag-of-words similarity above an adaptive threshold (starting at
0.5, backed off to 0.2 if too few negatives are available).  These
are genuinely well-matched pairs that share a similar title,
representing the exact scenario BoundaryHead must \emph{not} penalize.

\subsubsection{Training and Inference}

The head is trained with binary cross-entropy:
\begin{equation}
  \mathcal{L}_{\text{BCE}} =
    -\frac{1}{N}\sum_{i=1}^{N}
    \bigl[y_i \log \hat{p}_i + (1-y_i)\log(1-\hat{p}_i)\bigr],
  \label{eq:bce}
\end{equation}
where $\hat{p}_i = \sigma(\text{PairHead}(f(q_i), f(c_i)))$ and
$\sigma$ is the sigmoid function.  We use AdamW with learning rate
$10^{-3}$, weight decay $0.01$, batch size 128, and up to 40 epochs
with early stopping (patience~5).  Model selection is based on
validation loss.

At inference time, BoundaryHead score $s_{\text{boundary}}$ is
combined with the Stage-1 cosine score via Eq.~\eqref{eq:stage2}, with
$\lambda$ selected by grid search on the validation pool. The resulting system is therefore \textbf{mira-embeddings-v1}: an encoder-based reranking pipeline whose main gains come from progressive adaptation, plus a lightweight local correction head that suppresses residual role-boundary confusions without altering the underlying representation.

\paragraph{Boundary transfer diagnostic.}
BoundaryHead is trained on JD--JD pairs but applied to JD--CV pairs at inference.
A post-hoc diagnostic on 6,888 JD--CV pairs (199 JDs, Top-40 retrieved by
\textbf{mira-embeddings-v1-R2}, labels from Qwen3-32B) confirms that the boundary
signal transfers: $s_{\text{boundary}}$ achieves AUC~0.712 and Cohen's $d{=}0.791$
($p{<}0.0001$) in predicting LLM-judged boundary mismatches on unseen JD--CV pairs.
Full details are in the supplementary material.

\section{Evaluation}
\label{sec:eval}

\subsection{Research Questions and Evidence Map}
\label{sec:Questions}
We evaluate one primary claim and two supporting questions, each tied to a distinct source of evidence.

\textbf{RQ1 (primary claim).} Does recruitment-domain adaptation improve recall within the recruiter's actual review budget? This is the paper's central test, so our primary evidence comes from the \textbf{Local Pool}, where an upstream production retrieval stage first produces an initial shortlist and the model reranks the candidates that recruiters would actually inspect.

\textbf{RQ2 (mechanism).} If gains exist, where do they come from? We use staged ablations on the same Local Pool to separate JD-only adaptation, heterogeneous JD--CV adaptation, and boundary-aware reranking.

\textbf{RQ3 (supportive consistency).} Do the conclusions remain directionally consistent when the candidate pool extends beyond the initial shortlist? To examine this, we add a \textbf{supportive} larger-pool evaluation based on a union pool of 44,138 CVs with rubric-based LLM judgments. We treat this protocol as supporting evidence rather than definitive evidence because it covers only a subset of queries and does not replace full-corpus production evaluation.

\subsection{Experimental Setup}
\label{sec:setup}

\subsubsection{Local Pool: primary evaluation setting}
\label{sec:testpools}

We collect 300 real JDs from our company's internal recruitment data. For
each JD, an upstream production retrieval stage performs coarse
candidate generation and returns approximately 200 CVs.
Ground-truth positive candidates are pre-annotated per JD using
business-side labels. The 300 evaluation JDs are disjoint from all
source JDs used in synthetic-data generation and model training. The
evaluation task is to rerank these
${\sim}$200 initial candidates so that ground-truth positives
appear at the top. This setting simulates the actual production
pipeline, with initial candidate generation followed by embedding
reranking, and directly tests whether the specialized encoder and
BoundaryHead improve the coverage of qualified candidates within the
recruiter's review budget.

\subsubsection{Global Pool with LLM-as-Judge: supportive evaluation only}
\label{sec:judge}
\paragraph{Candidate pool and scoring setup.}
\label{sec:judge-setup}
To complement the production-style Local Pool, we construct a supportive larger-pool protocol to test whether the same directional conclusions hold when retrieval is performed directly over a broader corpus. Because exhaustive LLM judging over all 300 queries was prohibitively expensive, we pre-sampled \textbf{20 JDs} from the 300-JD test pool using stratified random sampling before model evaluation. The strata were defined by job family, seniority, and JD length, and sampling used a fixed random seed. We then unioned their attached candidates into a shared global pool of \textbf{44,138} unique CVs. Every model retrieves Top-200 candidates for each query JD directly from this corpus. We use this protocol as a supportive consistency check rather than as definitive evidence for full-query generalization. Additional judging details are provided in the supplementary material.

\paragraph{Scoring rubric.}
\label{sec:judge-rubric}
Qwen3-32B evaluates each JD--CV pair with a structured four-dimensional rubric (\textbf{Career}, \textbf{Skills}, \textbf{Education}, \textbf{Innovation}) and returns a weighted total score. A candidate is marked \textbf{judge-positive} when the total is at least \textbf{75}, with additional caps when JD MUST requirements are unmet or when only preferred requirements are missing. A condensed version of the judge prompt, together with the main cap rules and risk categories, is provided in the supplementary material.

\subsubsection{Baselines, metrics, and fair-comparison policy}
\label{sec:baselines}

\textbf{Baselines.}  We compare against eight general-purpose text
embedding models, all evaluated zero-shot without any domain
fine-tuning (Table~\ref{tab:baselines}). Jina Embeddings~v3 serves
simultaneously as the Baseline and as the backbone of mira-embeddings-v1.
Encoder-only variants
(e.g., mira-embeddings-v1-R1 or mira-embeddings-v1-R2 without BoundaryHead) are
treated as ablations rather than as the final proposed method.

\begin{table}[t]
  \caption{Compared models and approximate parameter counts. All
           non-proposed baselines are evaluated zero-shot.}
  \label{tab:baselines}
  \small
  \setlength{\tabcolsep}{4pt}
  \begin{tabular}{lcc}
    \toprule
    Model & Source & Params \\
    \midrule
    Jina Embeddings v3 (Baseline) & jinaai & 0.57B \\
    Jina Embeddings v4            & jinaai & 3.80B \\
    Jina Embeddings v5            & jinaai & 0.68B \\
    BGE-M3                        & BAAI & 0.57B \\
    GTE-Large-EN v1.5             & Alibaba-NLP & 0.31B \\
    Qwen3-Embedding-0.6B          & Qwen Team & 0.60B \\
    Qwen3-Embedding-4B            & Qwen Team & 4.00B \\
    Snowflake Arctic Embed L v2   & Snowflake & 0.57B \\
    mira-embeddings-v1 (Ours)    & Jina v3 + LoRA + BoundaryHead & 0.57B \\
    \bottomrule
  \end{tabular}
\end{table}

\textbf{Metrics.}  Following the recall-first framing of
Section~\ref{sec:problem}, we report Recall@$K$
(Eq.~\eqref{eq:recall}) as the primary metric and Precision@$K$
(Eq.~\eqref{eq:precision}) as a secondary guardrail, at
$K \in \{10, 20, 30, 40, 50, 60, 70\}$, macro-averaged over 300 JDs.
Precision@$K$ tracks whether recall improvements come at an
unacceptable cost to result quality: a model that meaningfully
sacrifices precision for recall is considered trade-off-unfavorable
within the recruiter review budget.

\textbf{Fair comparison.} In the Local Pool, all models rerank the same initial shortlist for each JD. In the Global Pool, all models retrieve under the same corpus of 44,138 CVs, the same Top-200 budget, and the same judge protocol. Because we do not include a recruitment-specific retriever baseline, our empirical claim is limited to improvement over strong general-purpose encoders in this production retrieval setting.

\subsection{Results and Analysis}
\subsubsection{Main result: does the method improve shortlist recall in the primary setting?}
\label{sec:comparison}

We begin with the \textbf{Local Pool}, which most closely matches the actual recruiter review scenario. Tables~\ref{tab:comparison-local-recall} and~\ref{tab:comparison-local-precision} report the full-system comparison against strong zero-shot general-purpose encoders.

\begin{table}[t]
  \caption{Local Pool heterogeneous text comparison on Recall@$K$. All values are percentages.}
  \label{tab:comparison-local-recall}
  \footnotesize
  \setlength{\tabcolsep}{3pt}
  \begin{tabularx}{\columnwidth}{@{}>{\raggedright\arraybackslash}p{0.24\columnwidth}*{7}{>{\centering\arraybackslash}X}@{}}
    \toprule
    Model & 10 & 20 & 30 & 40 & 50 & 60 & 70 \\
    \midrule
    Proposed & \textbf{30.75} & \textbf{52.10} & \textbf{63.49} & \textbf{73.74} & \textbf{77.55} & \textbf{83.91} & \textbf{88.00} \\
    GTE-Large & 26.14 & 43.72 & 60.47 & 68.87 & 74.82 & 78.98 & 84.16 \\
    Snowflake & 28.91 & 48.84 & 57.34 & 65.65 & 71.29 & 77.56 & 83.19 \\
    Qwen3-4B & 27.38 & 43.74 & 56.10 & 63.02 & 69.96 & 73.95 & 83.36 \\
    Jina5 & 28.40 & 43.97 & 60.05 & 68.50 & 73.27 & 77.62 & 81.36 \\
    Jina4 & 23.48 & 42.23 & 58.51 & 67.96 & 73.44 & 76.47 & 80.72 \\
    Jina v3 & 22.72 & 42.01 & 55.38 & 63.05 & 68.89 & 72.50 & 79.76 \\
    Qwen3-0.6B & 23.67 & 39.01 & 51.79 & 59.12 & 65.90 & 73.86 & 77.76 \\
    BGE-M3 & 23.24 & 32.24 & 41.51 & 51.37 & 56.58 & 61.43 & 67.85 \\
    \bottomrule
  \end{tabularx}
\end{table}

\begin{table}[t]
  \caption{Local Pool heterogeneous text  comparison on Precision@$K$. All
           values are percentages.}
  \label{tab:comparison-local-precision}
  \footnotesize
  \setlength{\tabcolsep}{3pt}
  \begin{tabularx}{\columnwidth}{@{}>{\raggedright\arraybackslash}p{0.24\columnwidth}*{7}{>{\centering\arraybackslash}X}@{}}
    \toprule
    Model & 10 & 20 & 30 & 40 & 50 & 60 & 70 \\
    \midrule
    Proposed & \textbf{39.62} & \textbf{34.81} & \textbf{31.92} & \textbf{29.33} & \textbf{26.19} & \textbf{24.33} & \textbf{22.33} \\
    GTE-Large & 35.38 & 31.54 & 28.72 & 25.67 & 23.73 & 21.96 & 20.97 \\
    Snowflake & 33.46 & 29.23 & 26.67 & 25.48 & 23.12 & 22.03 & 20.64 \\
    Qwen3-4B & 39.23 & 34.62 & 31.03 & 27.60 & 25.58 & 23.24 & 22.02 \\
    Jina5 & 35.77 & 31.54 & 29.49 & 27.12 & 24.12 & 22.09 & 20.64 \\
    Jina4 & 35.00 & 30.00 & 28.97 & 27.69 & 24.96 & 22.67 & 21.03 \\
    Jina v3 & 35.77 & 32.12 & 28.59 & 25.96 & 23.96 & 21.96 & 20.53 \\
    Qwen3-0.6B & 30.38 & 28.27 & 25.77 & 23.56 & 22.27 & 20.81 & 19.60 \\
    BGE-M3 & 26.54 & 23.08 & 20.77 & 20.00 & 19.04 & 18.12 & 17.24 \\
    \bottomrule
  \end{tabularx}
\end{table}

The proposed system achieves the highest Recall@$K$ at every evaluated cutoff, which is the dominant criterion in our recruiter-facing setting. For example, Recall@$10$, Recall@$50$, and Recall@$70$ reach 30.75\%, 77.55\%, and 88.00\%, all above the strongest zero-shot competitors. Despite its modest 0.6B size, the model outperforms much larger encoders such as Qwen3-Embedding-4B and Jina Embeddings~v4 on the primary recall metric. Precision also remains strong under the recall gain: mira-embeddings-v1 attains the highest Precision@$K$ at every reported cutoff, indicating that the improvement is not obtained by simply trading shortlist quality for broader coverage.

\subsubsection{Where do the gains come from? Staged ablation}
\label{sec:ablation}

We next examine which component drives the gain in the same \textbf{Local Pool} setting.

\begin{table}[t]
  \caption{Local Pool ablation results. R1/R2 abbreviate
           mira-embeddings-v1-R1/R2. Recall@$K$ is the primary
           criterion; Precision@$K$ serves as a deployment constraint.
           All values are percentages.}
  \label{tab:ablation-local}
  \scriptsize
  \resizebox{\columnwidth}{!}{%
  \begin{tabular}{c cccc cccc}
    \toprule
    & \multicolumn{4}{c}{\textbf{Recall@$K$ (\%)}} & \multicolumn{4}{c}{\textbf{Precision@$K$ (\%)}} \\
    \cmidrule(lr){2-5}\cmidrule(lr){6-9}
    $K$ & Baseline & R1 & R2 & Proposed & Baseline & R1 & R2 & Proposed \\
    \midrule
    10 & 22.72 & 26.99 & 30.10 & \textbf{30.75} & 35.77 & 30.77 & 38.85 & \textbf{39.62} \\
    20 & 42.01 & 42.96 & 50.74 & \textbf{52.10} & 32.12 & 28.08 & 34.23 & \textbf{34.81} \\
    30 & 55.38 & 56.29 & 61.70 & \textbf{63.49} & 28.59 & 25.77 & 30.77 & \textbf{31.92} \\
    40 & 63.05 & 65.40 & 70.45 & \textbf{73.74} & 25.96 & 23.56 & 27.88 & \textbf{29.33} \\
    50 & 68.89 & 70.38 & 77.36 & \textbf{77.55} & 23.96 & 21.19 & \textbf{26.19} & \textbf{26.19} \\
    60 & 72.50 & 75.06 & 83.77 & \textbf{83.91} & 21.96 & 19.59 & 24.21 & \textbf{24.33} \\
    70 & 79.76 & 80.12 & 87.85 & \textbf{88.00} & 20.53 & 18.89 & 22.24 & \textbf{22.33} \\
    \bottomrule
  \end{tabular}}
\end{table}

Table~\ref{tab:ablation-local} shows three clear patterns. First, \textbf{mira-embeddings-v1-R1} already improves recall over the zero-shot baseline at every cutoff, but the gain is not yet operationally stable: Precision@10 drops from 35.77\% to 30.77\%. This suggests that JD--JD adaptation improves semantic coverage, yet does not align the retrieval space tightly enough to preserve shortlist quality.

Second, \textbf{mira-embeddings-v1-R2} accounts for most of the system-level improvement. Once the encoder is further adapted with heterogeneous JD--CV training, recall rises sharply across all cutoffs, and precision also exceeds the baseline. Recall@50 improves from 68.89\% to 77.36\%, and Precision@10 increases from 35.77\% to 38.85\%. The second stage therefore does not merely broaden retrieval; it makes the representation more useful for practical recruiter reranking.

Third, \textbf{BoundaryHead} adds a smaller but consistent final gain on top of mira-embeddings-v1-R2. The full system achieves the best Recall@$K$ at every reported cutoff, with the clearest increment in the mid-range review budget where recruiters still inspect dozens rather than hundreds of candidates. Recall@30 improves from 61.70\% to 63.49\%, and Recall@40 improves from 70.45\% to 73.74\%. Precision remains flat or improves at nearly all cutoffs. Under a recall-first deployment objective, this makes \mbox{\textbf{mira-embeddings-v1}} the preferred default configuration.

\paragraph{Backbone generalizability.}
To confirm that the pipeline is not specific to Jina Embeddings~v3, we replicate the full R1$\to$R2$\to$
BoundaryHead sequence using Qwen3-Embedding-0.6B as the backbone.
Table~\ref{tab:backbone} reports results at selected cutoffs.
The same three-stage progression appears here as well: R1 yields a modest gain, R2 delivers the largest improvement, and BoundaryHead adds a further increment on both metrics.
The total pipeline lift on Recall@50 is $+8.19$\,pp, comparable to the $+8.66$\,pp gain on Jina v3, which suggests that the improvement pattern is backbone-agnostic.

\begin{table}[t]
  \caption{Backbone generalizability: Qwen3-Embedding-0.6B on the Local
           Pool across pipeline stages. Base = zero-shot; BH = BoundaryHead.}
  \label{tab:backbone}
  \scriptsize
  \resizebox{\columnwidth}{!}{%
  \begin{tabular}{c rrrr rrrr}
    \toprule
    & \multicolumn{4}{c}{Recall@$K$ (\%)} & \multicolumn{4}{c}{Precision@$K$ (\%)} \\
    \cmidrule(lr){2-5}\cmidrule(lr){6-9}
    $K$ & Base & $+$R1 & $+$R2 & $+$BH & Base & $+$R1 & $+$R2 & $+$BH \\
    \midrule
    10 & 23.67 & 25.58 & 26.59 & \textbf{27.67} & 30.38 & 33.08 & \textbf{34.23} & 33.46 \\
    30 & 51.79 & 52.60 & 56.86 & \textbf{57.69} & 25.77 & 26.79 & 27.82 & \textbf{28.97} \\
    50 & 65.90 & 66.85 & 71.68 & \textbf{74.09} & 22.27 & 23.12 & 23.50 & \textbf{24.35} \\
    70 & 77.76 & 80.31 & 83.57 & \textbf{85.57} & 19.60 & 20.15 & 21.08 & \textbf{21.36} \\
    \bottomrule
  \end{tabular}}
\end{table}

\subsubsection{Does the direction remain consistent beyond the initial shortlist?}
\label{sec:judge-results}

We finally turn to the broader \textbf{Global Pool} protocol as a supportive consistency check rather than as a replacement for the Local Pool.
\begin{table}[t]
  \caption{Global-pool ablation results under the LLM-as-Judge
           protocol. Recall@200 is the primary metric.}
  \label{tab:judge-ablation}
  \small
  \begin{tabular}{lcc}
    \toprule
    Model & Recall@200 & Precision@200 \\
    \midrule
    Baseline & 0.5969 & 0.7625 \\
    mira-embeddings-v1-R1 & 0.6066 & 0.6300 \\
    mira-embeddings-v1-R2 & 0.6634 & \textbf{0.8475} \\
    Proposed & \textbf{0.7047} & 0.8275 \\
    \bottomrule
  \end{tabular}
\end{table}

\begin{table}[t]
  \caption{Global-pool cross-model comparison under the LLM-as-Judge
           protocol.}
  \label{tab:judge-comparison}
  \small
  \begin{tabular}{lcc}
    \toprule
    Model & Recall@200 & Precision@200 \\
    \midrule
    Proposed & \textbf{0.7047} & 0.8275 \\
    Qwen3-4B & 0.6577 & \textbf{0.8550} \\
    Jina5 & 0.6971 & 0.8400 \\
    Qwen3-0.6B & 0.6619 & 0.8175 \\
    Baseline & 0.5969 & 0.7625 \\
    Snowflake v2 & 0.6356 & 0.7325 \\
    Jina4 & 0.6178 & 0.7275 \\
    GTE-Large-EN v1.5 & 0.6014 & 0.6300 \\
    BGE-M3 & 0.6454 & 0.5825 \\
    \bottomrule
  \end{tabular}
\end{table}

The broader-pool results follow the same directional pattern as the primary evaluation. In Table~\ref{tab:judge-ablation}, R1 remains a transitional stage, R2 provides the main gain, and the full system further improves Recall@200 from 0.6634 to 0.7047. In Table~\ref{tab:judge-comparison}, mira-embeddings-v1 achieves the highest Recall@200 among all evaluated encoders. At the same time, this supportive protocol also helps clarify the method's boundary: some larger zero-shot encoders achieve stronger single-point precision, and the final recall gain from BoundaryHead can come with a modest precision cost relative to R2. This fits our deployment target. The method is most useful when shortlist coverage is the dominant objective, rather than when a single-point precision target is optimized in isolation.



\section{Discussion}
Our experiments consistently show that a compact adaptation pipeline targeted at specific failure modes can outperform substantially larger general-purpose encoders across evaluation pools and backbone architectures. The more important implication is not simply the size of the recall gain, but what the result suggests about domain-specialized retrieval when the dominant errors are fine-grained rather than topical. Standard contrastive pretraining and scale-driven encoder upgrades largely assume that retrieval errors are spread across the similarity spectrum. In recruiter-facing shortlist generation, and plausibly in other high-stakes triage settings such as legal discovery and clinical matching, the errors that matter most to end-users appear instead to cluster in a narrow band of near-miss candidates that look plausible on surface features but fail on scope, seniority, or hard constraints. Once this concentration of errors is recognized, the practical response is not necessarily a uniformly stronger encoder, but a supervision design that explicitly identifies and suppresses the failure modes concentrated in that band. This reframing shifts the practical question from ``how much labeled data do we need'' to ``how precisely can we characterize the errors that matter.'' In settings where interaction labels are scarce but experts already understand the typical failure patterns, that is often a much cheaper bottleneck to address than large-scale annotation.

That said, this conclusion is bounded by several conditions. The entire pipeline assumes that role-boundary confusion is the dominant source of shortlist error. In markets or job families where other failure types dominate, such as temporal misalignment, credential misrepresentation, or regulatory eligibility, the current prompt pipeline would need to be redesigned around a different error taxonomy. Our evaluation is also tied to a single upstream retriever. A first-stage system with a substantially different recall--precision profile could change the residual error distribution enough to alter the marginal value of BoundaryHead. The Global Pool protocol likewise relies on a single LLM judge, and we have not yet validated its scoring against expert human annotators at scale, so judge-induced bias remains unquantified. These limitations point to several natural next steps: replacing the hand-designed prompt stages with an adaptive loop that mines dominant failure modes directly from first-stage retrieval errors, so supervision follows the observed error distribution rather than a fixed prior; extending the evaluation to multilingual and cross-domain triage tasks to test whether the failure-mode-targeted supervision principle transfers beyond Chinese-language recruitment; and selectively routing the boundary-confused subset identified by BoundaryHead to a lightweight cross-encoder pass to test how far the recall--precision frontier can be pushed without paying the cost of full pairwise rescoring.

\section{Conclusion}
\label{sec:conclusion}
In this work, we treat recruiter-facing JD-to-CV shortlist reranking as a recall-first problem and argue that its main bottleneck lies in role-boundary confusion rather than in generic semantic relevance alone. Our central finding is that boundary-targeted synthetic supervision, combined with progressive adaptation to recruitment semantics and the heterogeneous JD--CV gap, produces consistent recall gains over strong general-purpose embedding baselines in the primary local production-style pool, while a lightweight reranking head provides smaller local corrections for near-miss cases. These gains also do not appear to require costly industrial-scale interaction labels: a modest set of real JDs can be expanded with LLM-synthesized positives and hard negatives to create task-targeted supervision at low marginal cost. Evidence from the broader 44,138-candidate supplementary pool is directionally consistent with the local benchmark, but this auxiliary protocol is not a substitute for full-corpus production evidence. Our claim is therefore limited to recruiter-facing shortlist reranking under the evaluated setting, and it remains contingent on the quality of the synthesized supervision and the labeling rules used to construct the training signals.


\begin{acks}
Due to the open-source license terms of the Jina model used in this
work, the model is intended for research use only and not for
commercial deployment.
\end{acks}

\bibliographystyle{ACM-Reference-Format}
\bibliography{sample-base,paper}


\begin{thebibliography}{33}


\ifx \showCODEN    \undefined \def \showCODEN     #1{\unskip}     \fi
\ifx \showISBNx    \undefined \def \showISBNx     #1{\unskip}     \fi
\ifx \showISBNxiii \undefined \def \showISBNxiii  #1{\unskip}     \fi
\ifx \showISSN     \undefined \def \showISSN      #1{\unskip}     \fi
\ifx \showLCCN     \undefined \def \showLCCN      #1{\unskip}     \fi
\ifx \shownote     \undefined \def \shownote      #1{#1}          \fi
\ifx \showarticletitle \undefined \def \showarticletitle #1{#1}   \fi
\ifx \showURL      \undefined \def \showURL       {\relax}        \fi
\providecommand\bibfield[2]{#2}
\providecommand\bibinfo[2]{#2}
\providecommand\natexlab[1]{#1}
\providecommand\showeprint[2][]{arXiv:#2}

\bibitem[Bian et~al\mbox{.}(2019)]%
        {bian2019domain}
\bibfield{author}{\bibinfo{person}{Shuqing Bian}, \bibinfo{person}{Xu Zhao}, \bibinfo{person}{Biao Chang}, \bibinfo{person}{Kaimin Zhou}, \bibinfo{person}{Sheng Wen}, {and} \bibinfo{person}{Weidong Xiao}.} \bibinfo{year}{2019}\natexlab{}.
\newblock \showarticletitle{Domain Adaptation for Person-Job Fit with Transferable Deep Global Match Network}. In \bibinfo{booktitle}{\emph{Proceedings of the 2019 Conference on Empirical Methods in Natural Language Processing and the 9th International Joint Conference on Natural Language Processing (EMNLP-IJCNLP)}}. \bibinfo{pages}{4809--4819}.
\newblock
\href{https://doi.org/10.18653/v1/D19-1432}{doi:\nolinkurl{10.18653/v1/D19-1432}}


\bibitem[Bian et~al\mbox{.}(2020)]%
        {bian2020mvcn}
\bibfield{author}{\bibinfo{person}{Shuqing Bian}, \bibinfo{person}{Xu Zhao}, \bibinfo{person}{Biao Chang}, \bibinfo{person}{Kaimin Zhou}, \bibinfo{person}{Sheng Wen}, {and} \bibinfo{person}{Weidong Xiao}.} \bibinfo{year}{2020}\natexlab{}.
\newblock \showarticletitle{Learning to Match Jobs with Resumes from Sparse Interaction Data using Multi-View Co-Teaching Network}. In \bibinfo{booktitle}{\emph{Proceedings of the 29th ACM International Conference on Information and Knowledge Management (CIKM)}}. \bibinfo{pages}{65--74}.
\newblock
\href{https://doi.org/10.1145/3340531.3411929}{doi:\nolinkurl{10.1145/3340531.3411929}}


\bibitem[Bonifacio et~al\mbox{.}(2022)]%
        {bonifacio2022inpars}
\bibfield{author}{\bibinfo{person}{Luiz Bonifacio}, \bibinfo{person}{Hugo Abonizio}, \bibinfo{person}{Marzieh Fadaee}, {and} \bibinfo{person}{Rodrigo Nogueira}.} \bibinfo{year}{2022}\natexlab{}.
\newblock \showarticletitle{{InPars}: Unsupervised Dataset Generation for Information Retrieval}. In \bibinfo{booktitle}{\emph{Proceedings of the 45th International {ACM} {SIGIR} Conference on Research and Development in Information Retrieval}}. \bibinfo{pages}{2387--2392}.
\newblock
\href{https://doi.org/10.1145/3477495.3531863}{doi:\nolinkurl{10.1145/3477495.3531863}}


\bibitem[Chen et~al\mbox{.}(2024)]%
        {chen2024bgem3}
\bibfield{author}{\bibinfo{person}{Jianlv Chen}, \bibinfo{person}{Shitao Xiao}, \bibinfo{person}{Peitian Zhang}, \bibinfo{person}{Kun Luo}, \bibinfo{person}{Defu Lian}, {and} \bibinfo{person}{Zheng Liu}.} \bibinfo{year}{2024}\natexlab{}.
\newblock \showarticletitle{{BGE} {M3}-Embedding: Multi-Lingual, Multi-Functionality, Multi-Granularity Text Embeddings Through Self-Knowledge Distillation}.
\newblock \bibinfo{journal}{\emph{arXiv preprint arXiv:2402.03216}} (\bibinfo{year}{2024}).
\newblock


\bibitem[Dai et~al\mbox{.}(2023)]%
        {dai2023promptagator}
\bibfield{author}{\bibinfo{person}{Zhuyun Dai}, \bibinfo{person}{Vincent~Y. Zhao}, \bibinfo{person}{Ji Ma}, \bibinfo{person}{Yi Luan}, \bibinfo{person}{Jianmo Ni}, \bibinfo{person}{Jing Lu}, \bibinfo{person}{Anton Bakalov}, \bibinfo{person}{Kelvin Guu}, \bibinfo{person}{Keith~B. Hall}, {and} \bibinfo{person}{Ming{-}Wei Chang}.} \bibinfo{year}{2023}\natexlab{}.
\newblock \showarticletitle{Promptagator: Few-shot Dense Retrieval From 8 Examples}. In \bibinfo{booktitle}{\emph{The Eleventh International Conference on Learning Representations ({ICLR})}}.
\newblock
\urldef\tempurl%
\url{https://openreview.net/forum?id=gmL46YMpu2J}
\showURL{%
\tempurl}


\bibitem[Dettmers et~al\mbox{.}(2023)]%
        {dettmers2023qlora}
\bibfield{author}{\bibinfo{person}{Tim Dettmers}, \bibinfo{person}{Artidoro Pagnoni}, \bibinfo{person}{Ari Holtzman}, {and} \bibinfo{person}{Luke Zettlemoyer}.} \bibinfo{year}{2023}\natexlab{}.
\newblock \showarticletitle{{QLoRA}: Efficient Finetuning of Quantized Large Language Models}. In \bibinfo{booktitle}{\emph{Advances in Neural Information Processing Systems 36 ({NeurIPS})}}.
\newblock


\bibitem[Fang et~al\mbox{.}(2023)]%
        {fang2023recruitpro}
\bibfield{author}{\bibinfo{person}{Yuan Fang}, \bibinfo{person}{Lizi Liao}, \bibinfo{person}{Zhiyuan Zhao}, \bibinfo{person}{Dou Shen}, \bibinfo{person}{Min Zhang}, {and} \bibinfo{person}{Tat-Seng Chua}.} \bibinfo{year}{2023}\natexlab{}.
\newblock \showarticletitle{{RecruitPro}: A Pretrained Language Model with Skill-Aware Prompt Learning for Intelligent Recruitment}. In \bibinfo{booktitle}{\emph{Proceedings of the 29th ACM SIGKDD Conference on Knowledge Discovery and Data Mining (KDD)}}. \bibinfo{pages}{3991--4002}.
\newblock
\href{https://doi.org/10.1145/3580305.3599894}{doi:\nolinkurl{10.1145/3580305.3599894}}


\bibitem[Gao et~al\mbox{.}(2023)]%
        {gao2023hyde}
\bibfield{author}{\bibinfo{person}{Luyu Gao}, \bibinfo{person}{Xueguang Ma}, \bibinfo{person}{Jimmy Lin}, {and} \bibinfo{person}{Jamie Callan}.} \bibinfo{year}{2023}\natexlab{}.
\newblock \showarticletitle{Precise Zero-Shot Dense Retrieval without Relevance Labels}. In \bibinfo{booktitle}{\emph{Proceedings of the 61st Annual Meeting of the Association for Computational Linguistics (ACL)}}. \bibinfo{pages}{1762--1777}.
\newblock
\href{https://doi.org/10.18653/v1/2023.acl-long.99}{doi:\nolinkurl{10.18653/v1/2023.acl-long.99}}


\bibitem[Hu et~al\mbox{.}(2021)]%
        {hu2021lora}
\bibfield{author}{\bibinfo{person}{Edward~J. Hu}, \bibinfo{person}{Yelong Shen}, \bibinfo{person}{Phillip Wallis}, \bibinfo{person}{Zeyuan Allen-Zhu}, \bibinfo{person}{Yuanzhi Li}, \bibinfo{person}{Shean Wang}, \bibinfo{person}{Lu Wang}, {and} \bibinfo{person}{Weizhu Chen}.} \bibinfo{year}{2021}\natexlab{}.
\newblock \showarticletitle{{LoRA}: Low-Rank Adaptation of Large Language Models}.
\newblock \bibinfo{journal}{\emph{arXiv preprint arXiv:2106.09685}} (\bibinfo{year}{2021}).
\newblock


\bibitem[Izacard et~al\mbox{.}(2022)]%
        {izacard2022contriever}
\bibfield{author}{\bibinfo{person}{Gautier Izacard}, \bibinfo{person}{Mathilde Caron}, \bibinfo{person}{Lucas Hosseini}, \bibinfo{person}{Sebastian Riedel}, \bibinfo{person}{Piotr Bojanowski}, \bibinfo{person}{Armand Joulin}, {and} \bibinfo{person}{Edouard Grave}.} \bibinfo{year}{2022}\natexlab{}.
\newblock \showarticletitle{Unsupervised Dense Information Retrieval with Contrastive Learning}.
\newblock \bibinfo{journal}{\emph{Transactions on Machine Learning Research}} (\bibinfo{year}{2022}).
\newblock
\urldef\tempurl%
\url{https://openreview.net/forum?id=jKN1pXi7b0}
\showURL{%
\tempurl}


\bibitem[Jeronymo et~al\mbox{.}(2023)]%
        {jeronymo2023inparsv2}
\bibfield{author}{\bibinfo{person}{Vitor Jeronymo}, \bibinfo{person}{Luiz~Henrique Bonifacio}, \bibinfo{person}{Hugo Abonizio}, \bibinfo{person}{Marzieh Fadaee}, \bibinfo{person}{Roberto de~Alencar Lotufo}, \bibinfo{person}{Jakub Zavrel}, {and} \bibinfo{person}{Rodrigo Nogueira}.} \bibinfo{year}{2023}\natexlab{}.
\newblock \showarticletitle{{InPars-v2}: Large Language Models as Efficient Dataset Generators for Information Retrieval}.
\newblock \bibinfo{journal}{\emph{arXiv preprint arXiv:2301.01820}} (\bibinfo{year}{2023}).
\newblock


\bibitem[Jiang et~al\mbox{.}(2020)]%
        {jiang2020featurefusion}
\bibfield{author}{\bibinfo{person}{Chen Jiang}, \bibinfo{person}{Hengshu Zhu}, \bibinfo{person}{Wei Cheng}, \bibinfo{person}{Tong Xu}, \bibinfo{person}{Enhong Chen}, \bibinfo{person}{Hui Xiong}, {and} \bibinfo{person}{Le Liu}.} \bibinfo{year}{2020}\natexlab{}.
\newblock \showarticletitle{Learning Effective Representations for Person-Job Fit by Feature Fusion}. In \bibinfo{booktitle}{\emph{Proceedings of the 29th ACM International Conference on Information and Knowledge Management (CIKM)}}. \bibinfo{pages}{3005--3012}.
\newblock
\href{https://doi.org/10.1145/3340531.3412793}{doi:\nolinkurl{10.1145/3340531.3412793}}


\bibitem[Karpukhin et~al\mbox{.}(2020)]%
        {karpukhin2020dpr}
\bibfield{author}{\bibinfo{person}{Vladimir Karpukhin}, \bibinfo{person}{Barlas Oguz}, \bibinfo{person}{Sewon Min}, \bibinfo{person}{Patrick Lewis}, \bibinfo{person}{Ledell Wu}, \bibinfo{person}{Sergey Edunov}, \bibinfo{person}{Danqi Chen}, {and} \bibinfo{person}{Wen tau Yih}.} \bibinfo{year}{2020}\natexlab{}.
\newblock \showarticletitle{Dense Passage Retrieval for Open-Domain Question Answering}. In \bibinfo{booktitle}{\emph{Proceedings of the 2020 Conference on Empirical Methods in Natural Language Processing (EMNLP)}}. \bibinfo{pages}{6769--6781}.
\newblock
\href{https://doi.org/10.18653/v1/2020.emnlp-main.550}{doi:\nolinkurl{10.18653/v1/2020.emnlp-main.550}}


\bibitem[Khattab and Zaharia(2020)]%
        {khattab2020colbert}
\bibfield{author}{\bibinfo{person}{Omar Khattab} {and} \bibinfo{person}{Matei Zaharia}.} \bibinfo{year}{2020}\natexlab{}.
\newblock \showarticletitle{{ColBERT}: Efficient and Effective Passage Search via Contextualized Late Interaction over {BERT}}. In \bibinfo{booktitle}{\emph{Proceedings of the 43rd International {ACM} {SIGIR} Conference on Research and Development in Information Retrieval}}. \bibinfo{pages}{39--48}.
\newblock
\href{https://doi.org/10.1145/3397271.3401075}{doi:\nolinkurl{10.1145/3397271.3401075}}


\bibitem[Li et~al\mbox{.}(2023)]%
        {li2023gte}
\bibfield{author}{\bibinfo{person}{Zehan Li}, \bibinfo{person}{Xin Zhang}, \bibinfo{person}{Yanzhao Zhang}, \bibinfo{person}{Dingkun Long}, \bibinfo{person}{Pengjun Xie}, {and} \bibinfo{person}{Meishan Zhang}.} \bibinfo{year}{2023}\natexlab{}.
\newblock \showarticletitle{Towards General Text Embeddings with Multi-stage Contrastive Learning}.
\newblock \bibinfo{journal}{\emph{arXiv preprint arXiv:2308.03281}} (\bibinfo{year}{2023}).
\newblock


\bibitem[Ni et~al\mbox{.}(2022)]%
        {ni2022sentencet5}
\bibfield{author}{\bibinfo{person}{Jianmo Ni}, \bibinfo{person}{Gustavo Hern{\'a}ndez~{\'A}brego}, \bibinfo{person}{Noah Constant}, \bibinfo{person}{Ji Ma}, \bibinfo{person}{Keith~B. Hall}, \bibinfo{person}{Daniel Cer}, {and} \bibinfo{person}{Yinfei Yang}.} \bibinfo{year}{2022}\natexlab{}.
\newblock \showarticletitle{Sentence-{T5}: Scalable Sentence Encoders from Pre-trained Text-to-Text Models}. In \bibinfo{booktitle}{\emph{Findings of the Association for Computational Linguistics: {ACL} 2022}}. \bibinfo{pages}{1864--1874}.
\newblock
\href{https://doi.org/10.18653/v1/2022.findings-acl.146}{doi:\nolinkurl{10.18653/v1/2022.findings-acl.146}}


\bibitem[Nogueira and Cho(2019)]%
        {nogueira2019passage}
\bibfield{author}{\bibinfo{person}{Rodrigo Nogueira} {and} \bibinfo{person}{Kyunghyun Cho}.} \bibinfo{year}{2019}\natexlab{}.
\newblock \showarticletitle{Passage Re-ranking with {BERT}}.
\newblock \bibinfo{journal}{\emph{arXiv preprint arXiv:1901.04085}} (\bibinfo{year}{2019}).
\newblock


\bibitem[Nogueira et~al\mbox{.}(2020)]%
        {nogueira2020document}
\bibfield{author}{\bibinfo{person}{Rodrigo Nogueira}, \bibinfo{person}{Zhiying Jiang}, \bibinfo{person}{Ronak Pradeep}, {and} \bibinfo{person}{Jimmy Lin}.} \bibinfo{year}{2020}\natexlab{}.
\newblock \showarticletitle{Document Ranking with a Pretrained Sequence-to-Sequence Model}. In \bibinfo{booktitle}{\emph{Findings of the Association for Computational Linguistics: {EMNLP} 2020}}. \bibinfo{pages}{708--718}.
\newblock
\href{https://doi.org/10.18653/v1/2020.findings-emnlp.63}{doi:\nolinkurl{10.18653/v1/2020.findings-emnlp.63}}


\bibitem[Pradeep et~al\mbox{.}(2023)]%
        {pradeep2023rankzephyr}
\bibfield{author}{\bibinfo{person}{Ronak Pradeep}, \bibinfo{person}{Sahel Sharifymoghaddam}, {and} \bibinfo{person}{Jimmy Lin}.} \bibinfo{year}{2023}\natexlab{}.
\newblock \showarticletitle{{RankZephyr}: Effective and Robust Zero-Shot Listwise Reranking is a Breeze!}
\newblock \bibinfo{journal}{\emph{arXiv preprint arXiv:2312.02724}} (\bibinfo{year}{2023}).
\newblock


\bibitem[Qin et~al\mbox{.}(2018)]%
        {qin2018pjfnn}
\bibfield{author}{\bibinfo{person}{Chuan Qin}, \bibinfo{person}{Hengshu Zhu}, \bibinfo{person}{Tong Xu}, \bibinfo{person}{Chen Zhu}, \bibinfo{person}{Liang Jiang}, \bibinfo{person}{Enhong Chen}, {and} \bibinfo{person}{Hui Xiong}.} \bibinfo{year}{2018}\natexlab{}.
\newblock \showarticletitle{Enhancing Person-Job Fit for Talent Recruitment: An Ability-aware Neural Network Approach}. In \bibinfo{booktitle}{\emph{Proceedings of the 41st International {ACM} {SIGIR} Conference on Research and Development in Information Retrieval}}. \bibinfo{pages}{25--34}.
\newblock
\href{https://doi.org/10.1145/3209978.3210025}{doi:\nolinkurl{10.1145/3209978.3210025}}


\bibitem[Qin et~al\mbox{.}(2020)]%
        {qin2020enhanced}
\bibfield{author}{\bibinfo{person}{Chuan Qin}, \bibinfo{person}{Hengshu Zhu}, \bibinfo{person}{Tong Xu}, \bibinfo{person}{Chen Zhu}, \bibinfo{person}{Chao Ma}, \bibinfo{person}{Enhong Chen}, {and} \bibinfo{person}{Hui Xiong}.} \bibinfo{year}{2020}\natexlab{}.
\newblock \showarticletitle{An Enhanced Neural Network Approach to Person-Job Fit in Talent Recruitment}.
\newblock \bibinfo{journal}{\emph{ACM Transactions on Information Systems}} \bibinfo{volume}{38}, \bibinfo{number}{2} (\bibinfo{year}{2020}), \bibinfo{pages}{1--33}.
\newblock
\href{https://doi.org/10.1145/3376927}{doi:\nolinkurl{10.1145/3376927}}


\bibitem[Reimers and Gurevych(2019)]%
        {reimers2019sentencebert}
\bibfield{author}{\bibinfo{person}{Nils Reimers} {and} \bibinfo{person}{Iryna Gurevych}.} \bibinfo{year}{2019}\natexlab{}.
\newblock \showarticletitle{Sentence-{BERT}: Sentence Embeddings using {S}iamese {BERT}-Networks}. In \bibinfo{booktitle}{\emph{Proceedings of the 2019 Conference on Empirical Methods in Natural Language Processing and the 9th International Joint Conference on Natural Language Processing (EMNLP-IJCNLP)}}. \bibinfo{pages}{3982--3992}.
\newblock
\href{https://doi.org/10.18653/v1/D19-1410}{doi:\nolinkurl{10.18653/v1/D19-1410}}


\bibitem[Shen et~al\mbox{.}(2018)]%
        {shen2018joint}
\bibfield{author}{\bibinfo{person}{Dazhong Shen}, \bibinfo{person}{Hengshu Zhu}, \bibinfo{person}{Chen Zhu}, \bibinfo{person}{Tong Xu}, \bibinfo{person}{Chao Ma}, {and} \bibinfo{person}{Hui Xiong}.} \bibinfo{year}{2018}\natexlab{}.
\newblock \showarticletitle{A Joint Learning Approach to Intelligent Job Interview Assessment}. In \bibinfo{booktitle}{\emph{Proceedings of the Twenty-Seventh International Joint Conference on Artificial Intelligence ({IJCAI})}}. \bibinfo{pages}{3542--3548}.
\newblock
\href{https://doi.org/10.24963/ijcai.2018/492}{doi:\nolinkurl{10.24963/ijcai.2018/492}}


\bibitem[Sturua et~al\mbox{.}(2024)]%
        {sturua2024jinav3}
\bibfield{author}{\bibinfo{person}{Saba Sturua}, \bibinfo{person}{Isabelle Mohr}, \bibinfo{person}{Mohammad~Kalim Akram}, \bibinfo{person}{Michael G{\"u}nther}, \bibinfo{person}{Bo Wang}, \bibinfo{person}{Markus Krimmel}, \bibinfo{person}{Feng Wang}, \bibinfo{person}{Georgios Mastrapas}, \bibinfo{person}{Andreas Koukounas}, \bibinfo{person}{Nan Wang}, {and} \bibinfo{person}{Han Xiao}.} \bibinfo{year}{2024}\natexlab{}.
\newblock \showarticletitle{jina-embeddings-v3: Multilingual Embeddings With Task {LoRA}}.
\newblock \bibinfo{journal}{\emph{arXiv preprint arXiv:2409.10173}} (\bibinfo{year}{2024}).
\newblock


\bibitem[Su et~al\mbox{.}(2023)]%
        {su2023instructor}
\bibfield{author}{\bibinfo{person}{Hongjin Su}, \bibinfo{person}{Weijia Shi}, \bibinfo{person}{Jungo Kasai}, \bibinfo{person}{Yizhong Wang}, \bibinfo{person}{Yushi Hu}, \bibinfo{person}{Mari Ostendorf}, \bibinfo{person}{Wen{-}tau Yih}, \bibinfo{person}{Noah~A. Smith}, \bibinfo{person}{Luke Zettlemoyer}, {and} \bibinfo{person}{Tao Yu}.} \bibinfo{year}{2023}\natexlab{}.
\newblock \showarticletitle{One Embedder, Any Task: Instruction-Finetuned Text Embeddings}. In \bibinfo{booktitle}{\emph{Findings of the Association for Computational Linguistics: {ACL} 2023}}. \bibinfo{pages}{1102--1121}.
\newblock
\href{https://doi.org/10.18653/v1/2023.findings-acl.71}{doi:\nolinkurl{10.18653/v1/2023.findings-acl.71}}


\bibitem[Sun et~al\mbox{.}(2023)]%
        {sun2023rankgpt}
\bibfield{author}{\bibinfo{person}{Weiwei Sun}, \bibinfo{person}{Lingyong Yan}, \bibinfo{person}{Xinyu Ma}, \bibinfo{person}{Shuaiqiang Wang}, \bibinfo{person}{Pengjie Ren}, \bibinfo{person}{Zhumin Chen}, \bibinfo{person}{Dawei Yin}, {and} \bibinfo{person}{Zhaochun Ren}.} \bibinfo{year}{2023}\natexlab{}.
\newblock \showarticletitle{Is {ChatGPT} Good at Search? Investigating Large Language Models as Re-Ranking Agents}. In \bibinfo{booktitle}{\emph{Proceedings of the 2023 Conference on Empirical Methods in Natural Language Processing}}. \bibinfo{pages}{14918--14937}.
\newblock
\href{https://doi.org/10.18653/v1/2023.emnlp-main.923}{doi:\nolinkurl{10.18653/v1/2023.emnlp-main.923}}


\bibitem[Wang et~al\mbox{.}(2022)]%
        {wang2022e5}
\bibfield{author}{\bibinfo{person}{Liang Wang}, \bibinfo{person}{Nan Yang}, \bibinfo{person}{Xiaolong Huang}, \bibinfo{person}{Binxing Jiao}, \bibinfo{person}{Linjun Yang}, \bibinfo{person}{Daxin Jiang}, \bibinfo{person}{Rangan Majumder}, {and} \bibinfo{person}{Furu Wei}.} \bibinfo{year}{2022}\natexlab{}.
\newblock \showarticletitle{Text Embeddings by Weakly-Supervised Contrastive Pre-training}.
\newblock \bibinfo{journal}{\emph{arXiv preprint arXiv:2212.03533}} (\bibinfo{year}{2022}).
\newblock


\bibitem[Wang et~al\mbox{.}(2024)]%
        {wang2024improving}
\bibfield{author}{\bibinfo{person}{Liang Wang}, \bibinfo{person}{Nan Yang}, \bibinfo{person}{Xiaolong Huang}, \bibinfo{person}{Linjun Yang}, \bibinfo{person}{Rangan Majumder}, {and} \bibinfo{person}{Furu Wei}.} \bibinfo{year}{2024}\natexlab{}.
\newblock \showarticletitle{Improving Text Embeddings with Large Language Models}. In \bibinfo{booktitle}{\emph{Proceedings of the 62nd Annual Meeting of the Association for Computational Linguistics (Volume 1: Long Papers)}}. \bibinfo{pages}{11897--11916}.
\newblock
\href{https://doi.org/10.18653/v1/2024.acl-long.642}{doi:\nolinkurl{10.18653/v1/2024.acl-long.642}}


\bibitem[Xiao et~al\mbox{.}(2023)]%
        {xiao2023cpack}
\bibfield{author}{\bibinfo{person}{Shitao Xiao}, \bibinfo{person}{Zheng Liu}, \bibinfo{person}{Peitian Zhang}, \bibinfo{person}{Niklas Muennighoff}, \bibinfo{person}{Defu Lian}, {and} \bibinfo{person}{Jian{-}Yun Nie}.} \bibinfo{year}{2023}\natexlab{}.
\newblock \showarticletitle{{C-Pack}: Packaged Resources To Advance General Chinese Embedding}.
\newblock \bibinfo{journal}{\emph{arXiv preprint arXiv:2309.07597}} (\bibinfo{year}{2023}).
\newblock


\bibitem[Yan et~al\mbox{.}(2019)]%
        {yan2019interview}
\bibfield{author}{\bibinfo{person}{Rui Yan}, \bibinfo{person}{Ran Le}, \bibinfo{person}{Yang Song}, \bibinfo{person}{Tao Zhang}, \bibinfo{person}{Xiangliang Zhang}, {and} \bibinfo{person}{Dongyan Zhao}.} \bibinfo{year}{2019}\natexlab{}.
\newblock \showarticletitle{Interview Choice Reveals Your Preference on the Market: To Improve Job-Resume Matching through Profiling Memories}. In \bibinfo{booktitle}{\emph{Proceedings of the 25th {ACM} {SIGKDD} International Conference on Knowledge Discovery and Data Mining}}. \bibinfo{pages}{914--922}.
\newblock
\href{https://doi.org/10.1145/3292500.3330963}{doi:\nolinkurl{10.1145/3292500.3330963}}


\bibitem[Yu et~al\mbox{.}(2024)]%
        {yu2024confit}
\bibfield{author}{\bibinfo{person}{Zixuan Yu}, \bibinfo{person}{Qianlong Zhang}, {and} \bibinfo{person}{Ran Yu}.} \bibinfo{year}{2024}\natexlab{}.
\newblock \showarticletitle{{ConFit}: Improving Resume-Job Matching using Data Augmentation and Contrastive Learning}. In \bibinfo{booktitle}{\emph{Proceedings of the 18th ACM Conference on Recommender Systems (RecSys)}}. \bibinfo{pages}{282--291}.
\newblock
\href{https://doi.org/10.1145/3640457.3688133}{doi:\nolinkurl{10.1145/3640457.3688133}}


\bibitem[Yu et~al\mbox{.}(2025)]%
        {yu2025confitv2}
\bibfield{author}{\bibinfo{person}{Zixuan Yu}, \bibinfo{person}{Qianlong Zhang}, \bibinfo{person}{Ran Yu}, {and} \bibinfo{person}{Wolfgang Nejdl}.} \bibinfo{year}{2025}\natexlab{}.
\newblock \showarticletitle{{ConFit}~v2: Improving Resume-Job Matching using Hypothetical Resume Embedding and Runner-Up Hard-Negative Mining}. In \bibinfo{booktitle}{\emph{Findings of the Association for Computational Linguistics: ACL 2025}}.
\newblock


\bibitem[Zhu et~al\mbox{.}(2018)]%
        {zhu2018pjf}
\bibfield{author}{\bibinfo{person}{Chen Zhu}, \bibinfo{person}{Hengshu Zhu}, \bibinfo{person}{Hui Xiong}, \bibinfo{person}{Chao Ma}, \bibinfo{person}{Fang Xie}, \bibinfo{person}{Pengliang Ding}, {and} \bibinfo{person}{Pan Li}.} \bibinfo{year}{2018}\natexlab{}.
\newblock \showarticletitle{Person-Job Fit: Adapting the Right Talent for the Right Job with Joint Representation Learning}.
\newblock \bibinfo{journal}{\emph{ACM Transactions on Management Information Systems}} \bibinfo{volume}{9}, \bibinfo{number}{3} (\bibinfo{year}{2018}), \bibinfo{pages}{1--17}.
\newblock
\href{https://doi.org/10.1145/3234465}{doi:\nolinkurl{10.1145/3234465}}


\end{thebibliography}
\end{document}